\documentclass[runningheads]{llncs}
\usepackage[T1]{fontenc}
\usepackage{mathrsfs}
\usepackage{booktabs}
\usepackage{graphicx}
\usepackage{verbatim}
\usepackage{cite}
\usepackage[noend]{algorithm2e}
\usepackage{amsmath}
\usepackage{amssymb}
\usepackage{mathtools}
\usepackage{comment}  
\usepackage{wrapfig, multirow}
\usepackage{xcolor}
\usepackage{graphicx}
\usepackage{float}
\usepackage[textsize=tiny
            ]{todonotes}

\newcommand{\nk}[1]{\textcolor{black}{#1}}

\begin{document}

\nocite{*}

\title{On the Role of Domain Experts in Creating Effective Tutoring Systems}

\author{Sarath Sreedharan\inst{1}\orcidID{0000-0002-2299-0178} \and Kelsey Sikes\inst{1}\orcidID{0009-0003-9711-920X} 
 \and
Nathaniel Blanchard\inst{1}\orcidID{0000-0002-2653-0873} \and
Lisa Mason\inst{2}\orcidID{0009-0004-8847-5920}
\and 
Nikhil Krishnaswamy\inst{1}\orcidID{0000-0001-7878-7227}
\and
Jill Zarestky\inst{3}\orcidID{0000-0003-1728-1796}
}

\authorrunning{Sreedharan et al.}

\institute{
Department of Computer Science, Colorado State University, USA \\
\email{\{sarath.sreedharan, Kelsey.Sikes,  nathaniel.blanchard, nikhil.krishnaswamy\}@colostate.edu}
\and
Colorado State University Extension, USA\\
\email{lisa.mason@colostate.edu}
\and
School of Education, Colorado State University, USA \\
\email{jill.zarestky@colostate.edu}
}
%
%
%
\maketitle              

%
%
%
%

\begin{abstract}
The role that highly curated knowledge, provided by domain experts, could play in creating effective tutoring systems is often overlooked within the AI for education community. In this paper, we highlight this topic by discussing two ways such highly curated expert knowledge could help in creating novel educational systems. First, we will look at how one could use explainable AI (XAI) techniques to automatically create lessons. Most existing XAI methods are primarily aimed at debugging AI systems. However, we will discuss how one could use expert specified rules about solving specific problems along with novel XAI techniques to automatically generate lessons that could be provided to learners. Secondly, we will see how an expert specified curriculum for learning a target concept can help develop adaptive tutoring systems, that can not only provide a better learning experience, but could also allow us to use more efficient algorithms to create these systems. Finally, we will highlight the importance of such methods using a case study of creating a tutoring system for pollinator identification, where such knowledge could easily be elicited from experts.
\end{abstract}
\section{Introduction}

Artificial Intelligence (AI) as a field is currently in the midst of a transformative moment, marked by advances and achievements happening at a pace nearly unthinkable just a few years ago \cite{rossi_2025}.
Rapid advancement has also brought with it some implicit and explicit notions of how to build successful AI systems.
Chief among those notions is the push to downplay the need for, and in some cases completely reject, the use of any expert-specified knowledge \cite{sutton2019bitter}.
While AI for education, as a subfield, has always embraced the use of expert knowledge, we feel in this moment of the extreme proliferation of end-to-end trained AI systems; it is worth re-emphasizing the need and advantage of using expert-specified knowledge.
Through this position paper, we would like to go one step further and call for additional efforts to more effectively (a) gather, (b) represent, and (c) utilize experts in and for AI systems.

Our primary argument relates to the positioning of education in the larger landscape of possible AI applications and how it differs from the kind of problems on which the broader AI community typically focuses. To begin, many recent successes in AI relate to tacit knowledge tasks \cite{Kambhampati21}, like vision \cite{szeliski2022computer} and robotics \cite{ahn2022can}, where it is unclear how relevant expert knowledge could even be gathered. 
The other domain where there is a lot of recent excitement is the development and deployment of large-language models (LLM) \cite{bubeck2023sparks}.
These models operate on such large scales that gathering expert knowledge would never be cost-effective.
One could argue that LLM fine-tuning through processes like RLHF \cite{ouyang2022training} acts as a proxy for expert feedback.
However, these systems are able to get by with feedback from non-expert users, given that these systems are mostly deployed in low-stakes everyday use cases.

Education, on the other hand, can be a high-stakes application. For example, we do not want our systems to hallucinate and provide the learners with misleading or incorrect lessons; factually correct, appropriately scaffolded lessons that are responsive to individual learner needs are beyond the current capabilities of AI systems. 
While the state-of-the-art methods may not be able to provide the level of robustness and efficiency we would require for authentic or impactful educational settings~\cite{2024.EDM-posters.104}, there are ways we can accommodate the system's shortcomings by leveraging existing human expertise.
We have an extensive knowledge base on creating effective lessons and supporting learners across disciplinary domains (e.g. mathematics or history) and institutional contexts (e.g. elementary school or workplace learning). We also have experts who have the training to help craft approaches. While we want to minimize the demands placed on the experts, the solution should not be to ignore them. We should build systems that leverage expert input to help \nk{more learners with a wider variety of needs \cite{d2024learning}}.

To demonstrate our argument, we present two examples where expert-specified knowledge can be combined with state-of-the-art AI methods to generate better learning outcomes. First, we will show how explainable AI methods could be extended and augmented with expert-specified rules to generate lessons automatically. Second, we will examine how one could combine a hierarchically structured curriculum with a POMDP-based adaptive tutoring system to effectively generate lessons. \nk{To demonstrate the viability of these approaches, we ground them in an important citizen science learning problem of pollinator identification.}


Tracking pollinator health is a common goal shared by many community science (also known as citizen science) groups across the country \cite{oregonBee,mason_202AD,woodard2020towards}, particularly those groups that are concerned with issues of environmental sustainability or conservation. Such tracking can provide valuable ecological data pertaining to local environmental health. 
Learning more about insects and pollinators, including identification and behavior patterns, from experts is generally touted as one benefit of joining such groups.
It is also in the interest of group organizers to provide effective educational opportunities for these potential community scientists so they can carry out their scientific mission accurately and reliably.
Unfortunately, many of these organizations have capacity limitations, particularly around the availability of content domain experts, and do not have the resources to provide the level of individualized training some learners require.
The availability of effective automated instruction systems could make a huge impact on the ability to scale their organizational processes, recruit more participants, and guide those participants appropriately, ultimately enabling such organizations to achieve their scientific or environmental health aims \cite{schmeller2009advantages,kremen2011evaluating}.
As we will see in the following section, while such organizations may lack the resources to provide extensive training, they are fully capable of providing the level of information our proposed methods would require.
\section{Using Explainable AI Methods to Generate Lessons}
\begin{figure}
    \centering
    \includegraphics[width=0.8\linewidth]{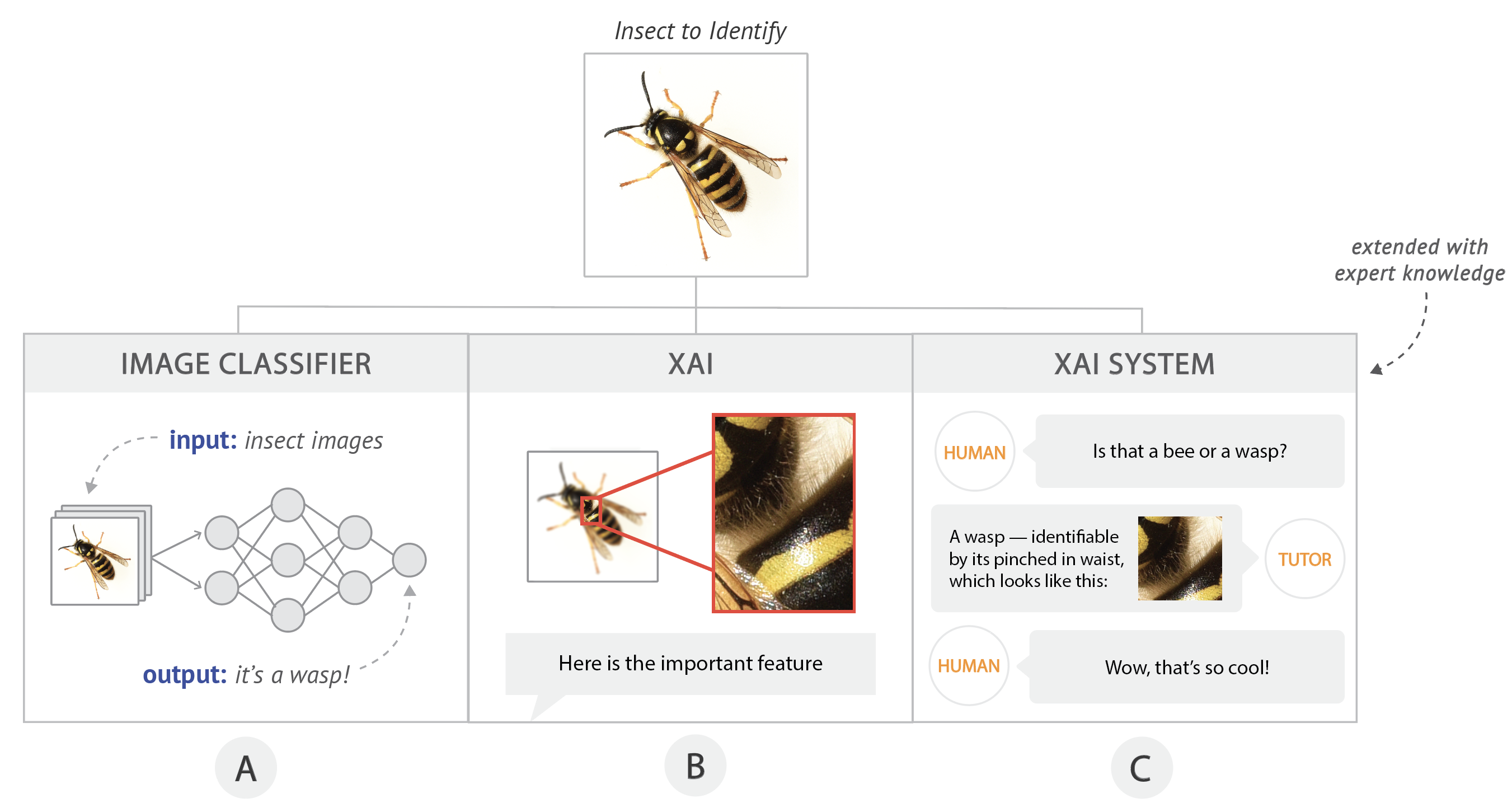}
    \caption{An example showing the application of our system.}
    \label{fig:enter-label}
\end{figure}
Explainable AI (XAI) as a field emerged out of the realization that as AI systems were becoming more capable, it was not always clear why the system was making the decisions it did \cite{gunning2017explainable}.
Even in a domain as specific as AI decision-making, the term explanation could include many aspects. 
However, most works in this space have focused on a few settings that the designers deemed most important.
Specifically, one could think of most works in this space as debugging mechanisms that could be used to verify whether the AI system is working the way it was expected.
These methods may be meant for the system designer to debug the algorithm/model or for the domain expert to check if it uses the correct problem features to make its decisions \cite{langley2019varieties}. To pick two different XAI techniques, consider feature attribution methods \cite{lundberg2017unified} and concept-based explanation methods \cite{kim2018interpretability}. In the former, the explanation includes highlighting parts of the input deemed most relevant to the decision. In the case of concept-based explanation, the explanation system can present high-level concepts that might influence the decision.

As a concrete example, consider the task of classifying a wasp presented in Figure \ref{fig:enter-label}. The subfigure (B) presents an example of a feature attribution method showing the image area that determined it was a wasp. A concept-based explanation might explain that the concept ``pinched-waist" influenced the decision. 
An entomologist can examine either explanation and determine that the system is correctly identifying the wasp using a relevant feature.

However, if we wanted to use this tool to help learners understand how to recognize a wasp, we quickly see that the two methods are insufficient.
If the learner is not aware of the concept of ``pinched-waist" or cannot visually recognize it, then neither of the explanations on its own will teach them about recognizing wasps.
On the other hand, these automatically generated pieces of information could be part of a pipeline to generate lessons for the learner.
We can create such lessons by using a curated list of rules to recognize wasps, expressed in terms of concepts that the XAI system can identify.
Once the AI system has identified an image to contain a wasp, one can iterate over the expert-specified rules provided and see which one can be surfaced to the learner and can be used to demonstrate the fact that the current image contains a wasp.
Subfigure (C) in Figure \ref{fig:enter-label} shows an example of how such a lesson might look for the given image.
Note that in this case, we can also account for cases where the AI system may be making incorrect predictions by associating uncertainty with the decision and the presence or absence of concepts.
Such approaches (cf. \cite{kim2018interpretability,sreedharanbridging}) have been previously used to ensure fidelity of generated explanation, in this case, is used to ensure the learner is getting accurate lessons.

\section{Expert-Specified Curriculum to Structure Lessons}
\begin{figure}
    \centering
    \includegraphics[width=0.75\linewidth]{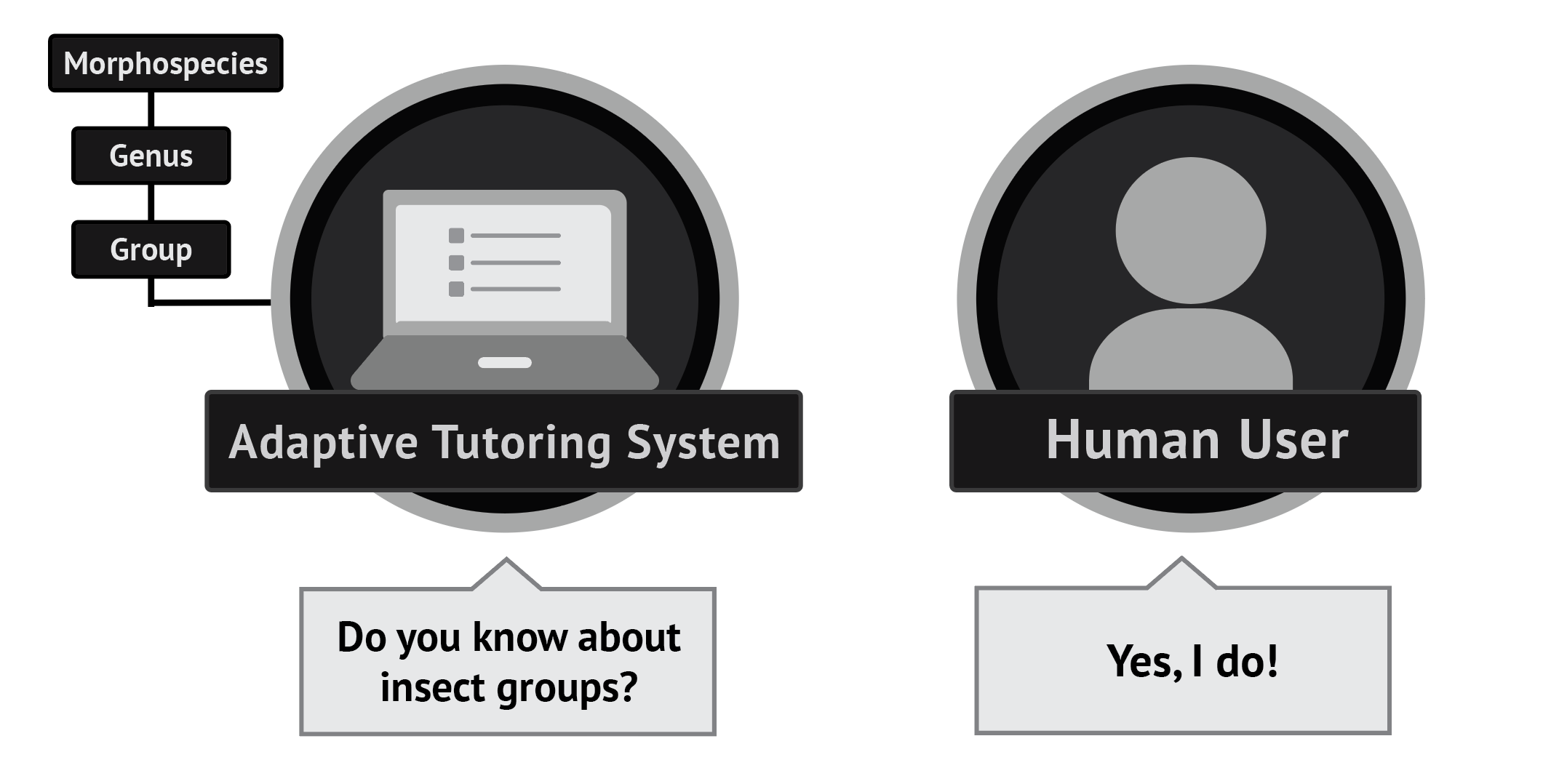}
    \caption{A hierarchical representation of concepts relevant to pollinator classification.}
    \label{fig:POMDP}
\end{figure}
Partially Observable Markov Decision Processes or POMDPs \cite{sondik1971optimal} have been widely discussed as a potential framework to represent adaptive tutoring systems \cite{rafferty2016faster}. They are particularly effective as they provide a natural way to encode the system's uncertainty about the learner's current knowledge. 
POMDPs also provide natural ways to capture actions to query the learner to test for their level of expertise and then provide lessons that might be better suited to it.

However, a naive application of POMDPs to most education settings remains infeasible as generating policies for them is quite computationally expensive \cite{murphy2000survey}. 
One of the biggest problems is the explosion of belief space that is needed to capture uncertainty related to the potential expertise the current learner might possess.
However, the use of a well-structured curriculum, created with the help of domain experts, could be helpful in simplifying the space. In particular, identifying hierarchies between concepts and the order in which they need to be taught could go a long way in simplifying the set of possible expertise levels the system needs to consider at any given point.

Going back to pollinator identification tasks, a curriculum could dictate that before they are trained to identify morphospecies, they should be able to identify the genus. Additionally, before they can recognize the genus, they should be able to recognize whether the given insect is a bee, wasp, or a fly \cite{mason2019assessing}. By encoding this order into the POMDP, you can have a policy that focuses on first testing the learner's ability to recognize the three groups before moving on to more granular distinction. 
Figure \ref{fig:POMDP} shows how a policy produced for such a POMDP might play out based on the feedback given by the learner.
This method of structuring belief spaces has shown to provide computational advantages in other settings \cite{sreedharan2021using}, and we expect it to carry over in this new setting as well.
\section{A Potential Tutoring System for Pollinator Identification}
\begin{figure}
    \centering
    \includegraphics[width=0.75\linewidth]{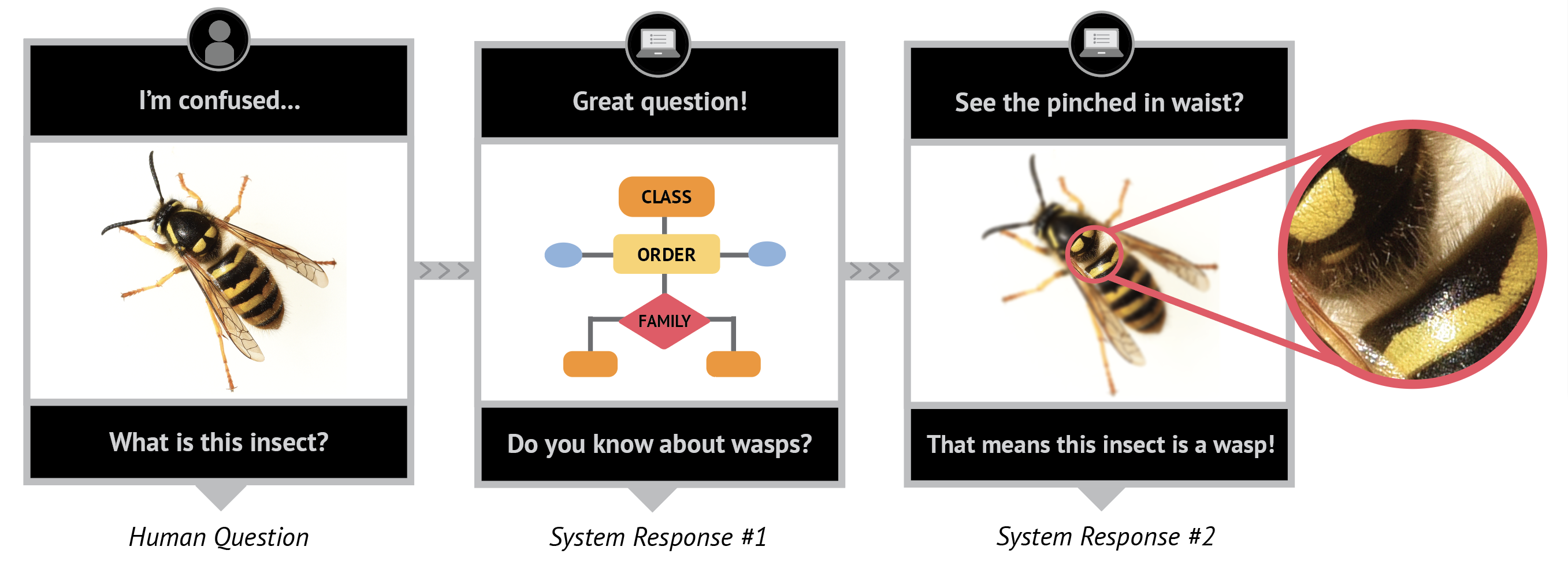}
    \caption{An overview of our proposed tutoring system.}
    \label{fig:overall}
\end{figure}

Putting these two pieces together, one can start seeing the outlines of an effective tutoring system that community scientists can use. To build such a system, we would need the domain experts to first build a curriculum that specifies in what order learners should be exposed to or master different concepts. Next, we would also require them to compile a set of rules that the learner can use to recognize different groups of insects.
While this may at first glance seem like a lot of information for an expert to provide, this is a set of information that is commonly provided to community scientists as part of their onboarding training.\footnote{Here are slides that were prepared for Native Bee Watch, a citizen science organization, by our co-author Lisa Mason \url{https://rb.gy/r03y0g}}
The morphospecies and concept classifiers needed to build the XAI pipeline that was described in the earlier section can be trained using data available from public repositories like iNaturalist \cite{nugent2018inaturalist}. One could also leverage the existing volunteer pool to get additional annotations and examples.
We can reuse the organization's existing teaching materials for the lessons and questions needed to build out the POMDP-based tutoring system. The objective function and the transition probabilities used by the POMDP can be initially set to default values and then refined over time based on learner feedback.

Once we have the two pieces in place, we can actually build a tutoring system that will allow the learners to take pictures of insects they see in the wild and ask the system to teach them about them. The POMDP-based system will query them to identify their current level of expertise and then determine the exact level of details that need to be included in the user feedback. This is then used to determine the rules that the XAI component needs to use to identify the feedback to be generated for the current image. The system can also provide them with more resources for them to better learn the target concept (in this case, a group, morphospecies).
However, building such a system with the current state-of-the-art AI methods would require the use of expert-specified knowledge. One that we argue is readily available in this case.
\section{Recommendations to the Community}
To conclude, here are our recommendations to the community.
\paragraph{Building Tools that Empower Experts.} Our first recommendation is to build better tooling and interfaces that will empower domain experts, who might not be AI experts, to easily specify their knowledge in forms that AI systems can use.  Potential early steps from the AI side in this direction include the use of symbolic interfaces \cite{kambhampati2022symbols} and the use of LLMs as semantic parsers \cite{guan2023leveraging}. It is also worth noting that the domain experts themselves may not be education experts, so there is a question about whether we can build tools that can help education experts introduce best practices into the knowledge we collect to support the integration of AI and domain expertise with high-quality education design.
\paragraph{Focusing on interdisciplinary teams.} While the AI for education community has always emphasized the importance of building interdisciplinary teams, this trend may erode as more researchers from the broader AI community engage with education as their target application. As a community, AI for education needs to emphasize building and supporting interdisciplinary teams.
\paragraph{Broader Public Education.} Finally, there is a need to educate the broader AI community, the public, and policymakers about what state-of-the-art AI systems can and cannot do in educational contexts. More work is needed to identify problem spaces where AI might be a valuable tool and where AI technology might still lag behind. We need to perform outreach activities to AI scientists so they better appreciate the challenges related to education and to the policymakers and public to make sure that resources are not being diverted to quixotic projects that are are bound to fail. Specifically, in the context of this proposal, we need to do a better job of identifying how expert knowledge can be best utilized and integrated at each level of education, including preK-12, higher ed, workforce development, and informal learning.

\begin{credits}
\subsection*{\ackname} This study was funded by NSF grant 2303019.
\end{credits}
%
%
%
%

\bibliographystyle{splncs04}

\bibliography{mybibliography}

\end{document}